\documentclass[11pt]{article}

\usepackage[final]{acl}

\usepackage{times}
\usepackage{latexsym}

\usepackage[T1]{fontenc}
\usepackage[utf8]{inputenc}

\usepackage{microtype}

\usepackage{inconsolata}

\usepackage{graphicx}

\usepackage{amsmath} % For \text and \emptyset
\usepackage{algorithm}
\usepackage{algpseudocode}
\usepackage{enumitem}
\usepackage{graphicx} % Required for inserting images
\usepackage{array}    % Required for better column formatting if needed
\usepackage{booktabs} % Optional: for better horizontal rules (alternative to \hline)
\usepackage{colortbl}
\usepackage{makecell}
\usepackage{blindtext}
\usepackage{bigstrut,multirow,rotating}
\definecolor{HighLight}{rgb}{0.96,0.92,0.96}
\newsavebox{\tablebox}
\title{Mixture of Reasonings: Teach Large Language Models to Reason with Adaptive Strategies}

\author{
 \textbf{Tao Xiong\textsuperscript{1, \dag}},
 \textbf{Xavier Hu\textsuperscript{2, \dag}},
 \textbf{Wenyan Fan\textsuperscript{2}},
 \textbf{Shengyu Zhang\textsuperscript{2, \ddag}}
 \\
 \\
 \textsuperscript{1}Dalian University of Technology,
 \textsuperscript{2}Zhejiang University
 \\
 \\
 \small{
\textbf{Correspondence:} 030130x@mail.dlut.edu.cn, sy\_zhang@zju.edu.cn
    }
 }

\begin{document}
\maketitle
\begin{abstract}
Large language models (LLMs) excel in complex tasks through advanced prompting techniques like Chain-of-Thought (CoT) and Tree-of-Thought (ToT), but their reliance on manually crafted, task-specific prompts limits adaptability and efficiency. 
% We introduce Mixture of Reasoning (MoR), a novel training framework that embeds diverse reasoning strategies into LLMs, enabling autonomous, task-adaptive reasoning without external prompt engineering. MoR operates in two phases: Thought Generation, creating large-scale reasoning chain templates using advanced models like GPT-4o, and SFT Dataset Construction, pairing these templates with benchmark datasets for supervised fine-tuning. 
We introduce Mixture of Reasoning (MoR), a training framework that embeds diverse reasoning strategies into LLMs for autonomous, task-adaptive reasoning without external prompt engineering. MoR has two phases: Thought Generation, creating reasoning chain templates with models like GPT-4o, and SFT Dataset Construction, pairing templates with benchmark datasets for supervised fine-tuning.
Our experiments show that MoR significantly enhances performance, with $MoR_{150}$ achieving 0.730 (2.2\% improvement) using CoT prompting and 0.734 (13.5\% improvement) compared to baselines. MoR eliminates the need for task-specific prompts, offering a generalizable solution for robust reasoning across diverse tasks.
\end{abstract}

\renewcommand{\thefootnote}{}
\footnotetext{$^{\dag}$Both authors contributed equally to this research.\\ $^{\ddag}$Corresponding Author} 
\renewcommand{\thefootnote}{\arabic{footnote}}

\section{Introduction}
Large language models (LLMs) have achieved remarkable success across diverse domains, largely due to advanced prompting techniques such as Chain-of-Thought (CoT) \citep{wei2023chainofthoughtpromptingelicitsreasoning} , Tree-of-Thought (ToT) \citep{yao2023treethoughtsdeliberateproblem}, and Prompt-of-Thought (PoT) \citep{10825021}. These methods guide models to reason step-by-step or explore multiple reasoning paths, significantly enhancing their performance on complex tasks. However, their effectiveness heavily relies on manually crafted, task-specific prompts, which are time-consuming to design and challenging to adapt optimally across varied tasks. This dependency on prompt engineering poses a critical bottleneck, where generic prompts often fail to elicit robust reasoning.

To address this challenge, we propose Mixture of Reasoning (MoR), a novel training framework that embeds a diverse set of reasoning strategies directly into LLMs, enabling them to autonomously select and apply effective reasoning methods tailored to specific tasks. Unlike existing approaches \citep{gao2024metareasoninglargelanguage, zhou2024selfdiscoverlargelanguagemodels} that rely on external prompt engineering to elicit reasoning, MoR internalizes reasoning capabilities by fine-tuning models on a curated supervised fine-tuning (SFT) dataset enriched with reasoning chain templates. These templates, generated by leveraging the advanced reasoning abilities of closed-source large models (e.g., GPT-4o), cover a wide range of reasoning patterns, including multi-step deduction, analogical reasoning, and strategic thinking. The MoR framework operates in two key phases: (1) Thought Generation, where we produce large-scale reasoning chain templates (e.g., 50, 150, 300, and 500 chains) to capture diverse problem-solving approaches, and (2) SFT Dataset Construction, where we pair these templates with samples from benchmark datasets to create a training dataset that teaches models to adaptively apply reasoning strategies. By embedding these strategies into the model’s parameters, MoR eliminates the need for task-specific prompt design and enhances generalizability across complex reasoning tasks. 

Our experiments demonstrate that MoR significantly outperforms baseline models, with our best model, $MoR_{150}$, achieving a performance of 0.730 with CoT prompting (a 2.2\% improvement over the baseline) and 0.734 with direct IO prompting (a 13.5\% improvement), showcasing its ability to reason effectively without explicit guidance.

Our contributions are as follows:
\begin{itemize}
    \item We introduce MoR, a training framework that embeds diverse reasoning strategies into LLMs, enabling task-adaptive reasoning without reliance on specific prompts.
    \item We propose a two-step methodology involving Thought Generation and SFT Dataset Construction, leveraging large-scale reasoning templates and curated datasets.
    \item We provide comprehensive experimental evidence demonstrating MoR’s superiority over baseline models, with detailed analyses and case studies illustrating its logical reasoning capabilities.
\end{itemize}
\begin{figure*}[!t]
    \centering
    \includegraphics[width=1\linewidth]{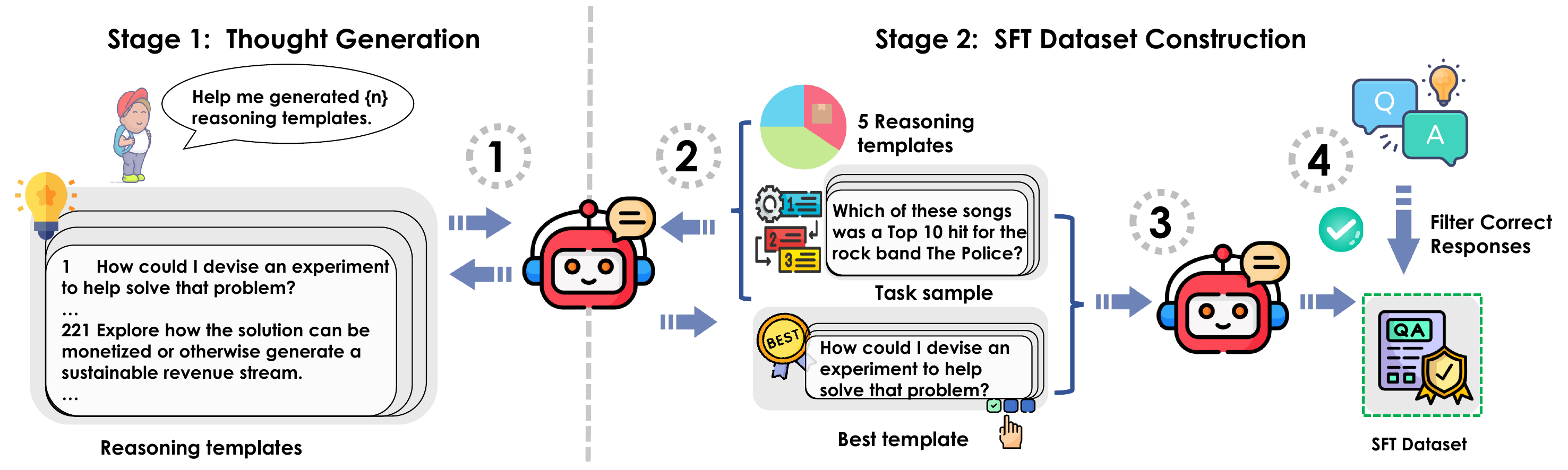}
    \caption{Overview of our proposed MoR framework. The MoR framework can be divided into two stages: (1) Thought Generation. As shown in step 1, this involves generating a large number of reasoning chain templates using GPT. (2) SFT Dataset Construction. As depicted in steps 2, 3, and 4, this includes selecting optimal reasoning chains, creating prompts, and filtering for correct responses.}
    \label{fig:xot_framework}
\end{figure*}

% XoT method可细分为两个阶段：（1）X-chain of Thought Generation. 如序号1所示，通过GPT生成大量的思维链模板。（2）SFT Dataset Construction. 如序号2，3，4所示，by selecting optimal reasoning chains, creating prompts, and filtering for correct responses.
% The process begins with the generation of reasoning templates through the X-chain of Thought, followed by the construction of the Supervised Fine-Tuning (SFT) dataset by selecting optimal reasoning chains, creating prompts, and filtering for correct responses.

\section{Related Work}
\textbf{Supervised Fine-Tuning of Large Language Models.} Supervised Fine-Tuning (SFT) \citep{zhang2024instructiontuninglargelanguage} leverages structured (instruction-answer) pairs to fully exploit the zero-shot capabilities of large models. This process enables models to learn systematic reasoning patterns and produce accurate results on complex reasoning tasks. By fine-tuning on task-specific datasets, SFT emphasizes the development of logical reasoning, problem-solving skills, and domain-specific knowledge. In recent years, numerous studies on SFT for large models have emerged, including approaches such as zeroth-order fine-tuning \citep{malladi2024finetuninglanguagemodelsjust} and robust fine-tuning \citep{tian2023fasttrainableprojectionrobust}. Notably, SFT has demonstrated significant advantages in reasoning-related fields, particularly in mathematics \citep{cobbe2021trainingverifierssolvemath, chen2024autoprmautomatingproceduralsupervision} and code generation \citep{wang2024dolphcoderecholocatingcodelarge}, achieving promising results.

\textbf{Prompt Engineering.} Thoughtful prompt design can enhance the reasoning abilities of large models, helping them tackle complex challenges. Chain-of-thought prompting is a strategy that guides large language models (LLMs) to produce intermediate reasoning steps, ultimately leading to the final answer and improving problem-solving accuracy. Typical implementations include zero-shot CoT \citep{kojima2023largelanguagemodelszeroshot} and few-shot CoT \citep{wei2023chainofthoughtpromptingelicitsreasoning}. Recent studies \citep{yasunaga2024largelanguagemodelsanalogical,zheng2024stepbackevokingreasoning, wang2024unleashingemergentcognitivesynergy, wilf2023thinktwiceperspectivetakingimproves} have further advanced this method by integrating more structured algorithms and search strategies. For example, \citet{zheng2024stepbackevokingreasoning} enables LLMs to abstract high-level concepts and first principles from detailed instances, while \citet{yasunaga2024largelanguagemodelsanalogical} prompts models to generate relevant examples or contextual knowledge before solving the problem. Additionally, some research \citep{gao2024metareasoninglargelanguage, zhou2024selfdiscoverlargelanguagemodels} is also exploring the use of different types of reasoning chains tailored to various task categories. Our approach, MoR, differs from these methods in that it not only produces a diverse array of reasoning strategies but also employs supervised fine-tuning (SFT) to train a foundational model capable of multi-chain reasoning.

\section{Method}
\label{sec:method}
In this section, we will provide a detailed description of the specific implementation of the MoR method. \textbf{\textit{The framework is shown in Figure \ref{fig:xot_framework}}}, and we have divided the MoR method into two steps: (1) Thought Generation: generating multiple thought chains to expand the model's thinking approach. (2) SFT Dataset Construction: creating an SFT training dataset using various thinking approaches.

\subsection{Thought Generation}
\label{sec:X_chain_of_thought_generation}
For small parameter models, due to insufficient embedded knowledge and limited reasoning capabilities, simply instructing them with "Let's think step by step" does not effectively stimulate the model's capabilities.

To address this issue, we first need to provide the model with effective thinking approaches for different types of problems. Existing methods \citep{wei2023chainofthoughtpromptingelicitsreasoning,yasunaga2024largelanguagemodelsanalogical,zheng2024stepbackevokingreasoning} mainly focus on generating specific thinking approaches for one type of problem. We decided to leverage the reasoning ability of closed-source large models. Initially, we prompted GPT to generate a large number of reasoning chain templates for reasoning tasks. In this section, we pre-generated \underline{\textbf{50}}, \underline{\textbf{150}}, \underline{\textbf{300}}, and \underline{\textbf{500}} reasoning chains, denoted as $T = {t_1, t_2, ..., t_M}$.

\subsection{SFT Dataset Construction}

After generating the reasoning chains in \S \ref{sec:X_chain_of_thought_generation}, we need to construct an MoR dataset that can be used for training. In this section, we select several commonly used reasoning datasets, such as HotpotQA, StrategyQA, MMLU, BigTom, and Trivial Creative Writing (more details will be discussed in \S \ref{sec:setup}).

First, we randomly select Specified quantity samples \textbf{N} from each dataset as training samples. Then, for the selected dataset $D_{\text{source}} = \{s_1, s_2, \dots, s_K\}$, where K=N, we randomly select 5 reasoning chain templates $T_{\text{sub}}$ from the reasoning chain template set $T = \{t_1, t_2, \dots, t_M\}$, forming a subset of reasoning chains. The selected samples $D_{\text{selected}}$ along with the selected subset are then fed into GPT, which selects the reasoning chain $T_{\text{best}}$ it deems most beneficial for solving the problem based on the problem structure of the samples.
Next, we create a prompt by combining the selected reasoning chain template $T_{\text{best}}$ with each sample $s_i$, and feed it to the model for reasoning. After evaluation, we filter out the correct answers, and the resulting set is combined into an SFT dataset $D_{\text{SFT}}$.

% \subsection{Model Fine-Tuning}
% xxx

\begin{algorithm}[H] % H forces the algorithm to be placed "here"
\caption{SFT Dataset Construction}
\label{alg:sft_construction}
\begin{algorithmic}[1] % The [1] enables line numbering
    \State $D_{\text{SFT}} \gets \emptyset$
    % \For{$i \gets 1$ to $N$}
    \For{$i \gets 1$ to $N$}
        \State $s_i \gets D_{\text{selected}}[i]$ // Get the $i$-th sample
        \State $T_{\text{sub}} \gets \text{RandomSelect}(T, N)$ // Select $N$ templates 
        \State $Prompt_{\text{select}} \gets \text{FormatSelectPrompt}(s_i, T_{\text{sub}})$ 
        \State $t_{\text{best}} \gets \texttt{LLM}.\text{infer}(Prompt_{\text{select}})$ 
        \State $Prompt_{\text{reason}} \gets \text{FormatReasonPrompt}(s_i, t_{\text{best}})$ 
        \State $R_i \gets \text{model}.\text{infer}(Prompt_{\text{reason}})$ 
        \State $IsCorrect \gets \text{Eval}(s_i, R_i)$ //Evaluate if $R_i$ is correct for $s_i$
        \If{$IsCorrect$ is True}
            \State $SFT_{\text{entry}} \gets \text{FormatForSFT}(s_i, R_i)$ 
            \State $D_{\text{SFT}} \gets D_{\text{SFT}} \cup \{SFT_{\text{entry}}\}$ 
        \EndIf
    \EndFor
    \State \Return $D_{\text{SFT}}$ //Return the constructed SFT dataset
\end{algorithmic}
\end{algorithm}

\begin{table*}[ht]
\centering
\footnotesize % \footnotesize\normalsize
\renewcommand{\arraystretch}{1.1} % 调整行高

% 先缩小列宽
\begin{lrbox}{\tablebox}
\begin{tabular}{
  >{\centering\arraybackslash}p{1.8cm}  % Model列宽
  >{\centering\arraybackslash}p{1.2cm}| % Prompt列宽
  >{\centering\arraybackslash}p{1.0cm}  % 其余列宽依次缩小
  >{\centering\arraybackslash}p{1.2cm}
  >{\centering\arraybackslash}p{0.8cm}
  >{\centering\arraybackslash}p{0.9cm}
  >{\centering\arraybackslash}p{2.0cm}
  >{\centering\arraybackslash}p{0.8cm}
}
\hline
\multicolumn{2}{c|}{\textbf{Method}} &
\multirow{2}{*}{\textbf{Hotpotqa}} &
\multirow{2}{*}{\makecell{\textbf{Strate-}\\ \textbf{gyqa}}} &
\multirow{2}{*}{\textbf{MMLU}} &
\multirow{2}{*}{\textbf{BigTom}} &
\multirow{2}{*}{\makecell{\textbf{Trivial Creative} \\ \textbf{writing}}} &
\multirow{2}{*}{\textbf{overall}} \\
\cmidrule(lr){1-2}
\textbf{Model} & \textbf{Prompt} &  &  &  &  &  & \\
\hline
\hline

\multirow{2}{*}{Qwen2.5-7B} 
    & IO  & 1.00 & 0.400 & 0.540 & 0.688 & 0.368 & 0.599 \\
    & CoT & 0.980 & 0.940 & 0.560 & 0.750 & 0.308 & \textbf{0.708} \\
\hline

% \multirow{2}{*}{$MoR_{49}$} 
%     & IO & 0.540 & 0.920 & 0.700 & 0.863 & 0.276 & \textbf{0.660} \\
%     & CoT  & 0.780    & 0.820    & 0.620    & 0.825    & 0.236    & 0.656 \\
% \hline

\multirow{2}{*}{$MoR_{50}$} 
    & IO & 0.540 & 0.900 & 0.580 & 0.888 & 0.336 & \textbf{0.649} \\
    & CoT  & 0.640    & 0.480    & 0.580    & 0.925    & 0.300    & 0.585 \\
\hline

\multirow{2}{*}{$MoR_{150}$} 
    & IO & 0.98 & 0.94 & 0.560 & 0.875 & 0.144 & 0.700 \\ % 0.6998
    & CoT  & 0.98    & 0.920    & 0.620    & 0.900    & 0.232    & \textbf{0.730} \\ % 0.7304
\hline

\multirow{2}{*}{$MoR_{300}$} 
    & IO & 0.980 & 0.840 & 0.480 & 0.938 & 0.208 & 0.689 \\
    & CoT  & 0.980    & 0.880    & 0.560    & 0.863    & 0.292    & \textbf{0.715} \\
\hline

\multirow{2}{*}{$MoR_{500}$} 
    & IO & 0.960 & 0.920 & 0.620 & 0.913 & 0.256 & \textbf{0.734} \\
    & CoT  & 0.960    & 0.900    & 0.500    & 0.900    & 0.276    & 0.707 \\
\hline
\hline
\multirow{2}{*}{\makecell{Qwen2.5-7B \\ (Expend)}} 
    & IO & 0.960 & 0.400 & 0.595 & 0.731 & 0.368 & 0.611 \\
    & CoT  & 0.915  & 0.885  & 0.565  & 0.738  & 0.308  & \textbf{0.682} \\
\hline
\multirow{2}{*}{$MoR_{150}$(Expend)} 
    & IO & 0.990 & 0.880 & 0.610 & 0.863 & 0.144 & 0.697 \\
    & CoT  & 0.960    & 0.905    & 0.600    & 0.919    & 0.232    & \textbf{0.723} \\
\hline
\end{tabular}
\end{lrbox}

% 整体缩放表格到单栏宽度·
\resizebox{\linewidth}{!}{\usebox{\tablebox}}

\caption{Performance on reasoning tasks. We selected Qwen2.5-7B-instruct as the baseline model. We train the baseline model using our MoR approach by varying the number of reasoning chain templates. Additionally, to evaluate the effectiveness of MoR, we extend the test set from 50 to 200 instances, specifically comparing the baseline model with $MoR_{150}$. The best results for each setting are bolded.}
\label{tab:MoR_result}
\end{table*}

% Performance on reasoning tasks. 我们使用MoR方法通过按照不同思维链模板数量在基线模型Qwen2.5-7B-INS模型上进行训练。同时，针对基线模型与$MoR_(150)$，我们扩展测试集到200，进一步确保MoR方法的有效性。
% 基线模型我们选取Qwen2.5-7B-instruct，We selected Qwen2.5-7B-instruct as the baseline model. The best results for each configuration are bolded.
\begin{figure*}[!t]
    \centering
    \includegraphics[width=1\linewidth]{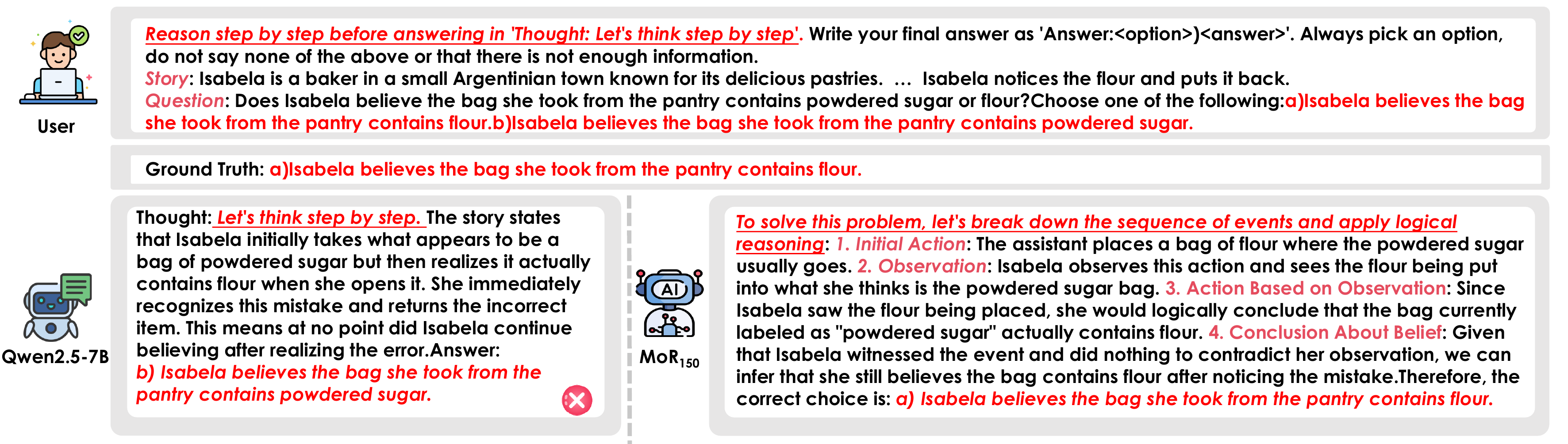}
    \caption{Case study comparing the baseline model and $MoR_{150}$ using CoT prompts. The Qwen2.5-7B-instruct model follows the "Let's think step by step." approach but ultimately produces incorrect answers. In contrast, the $MoR_{150}$ model adopts the MoR reasoning method, analyzing problems logically and ultimately arriving at the correct answer.}
    \label{fig:case_study}
\end{figure*}
\section{Experiment}
\subsection{Setup}
\label{sec:setup}
\textbf{Datasets.} 
% In the experiment, we selected five reasoning datasets, and below is a detailed introduction to each. For the testing set, we randomly chose 50 samples from each dataset. Specifically, for BigTom, we selected 20 samples across four different "belief settings," totaling 80 samples. Specifically, during the construction of the SFT dataset, as mentioned in \S \ref{sec:method}, we used the GPT-4o-2024-08-06 version of GPT.
In the experiment, we selected five reasoning datasets, with 50 samples randomly chosen from each dataset for testing. For BigTom, we selected 20 samples across four different "belief settings," totaling 80 samples. The SFT dataset construction used the GPT-4o-2024-08-06 version of GPT, as mentioned in \S \ref{sec:method}.
\begin{itemize}[leftmargin=*]
    \item {
            \textbf{HotpotQA} \citep{yang2018hotpotqadatasetdiverseexplainable}: %This dataset is designed for question answering, featuring complex, multi-hop questions. It includes strong supervision by providing supporting facts, enabling the development of more interpretable question answering systems.
            HotpotQA is designed for question answering with complex, multi-hop questions and strong supervision for interpretable systems. 
        }
    \item {
       \textbf{StrategyQA} \citep{geva2021didaristotleuselaptop}: %StrategyQA introduces a question answering benchmark where the reasoning steps needed to answer the question are not explicitly provided but must be inferred through strategic thinking.
       StrategyQA requiring inference of reasoning steps for question answering through strategic thinking.
    }
    
    \item {
         \textbf{MMLU} \citep{hendrycks2021measuringmassivemultitasklanguage}: MMLU is an extensive multitask benchmark composed of multiple-choice questions across a wide range of knowledge domains. The benchmark spans 57 subjects across diverse domains.
         %, including humanities, social sciences, natural sciences, mathematics, US history, computer science, and law.
         % A multitask benchmark with 57 subjects across diverse knowledge domains. 
        %subjects such as humanities, social sciences, natural sciences, and other fields crucial for learning, with 57 tasks spanning areas like elementary mathematics, US history, computer science, law, and more.
    }
    
    \item {
       \textbf{BigTom} \citep{wilf2023thinktwiceperspectivetakingimproves}: BigTom is a benchmark for assessing the Theory of Mind (ToM) reasoning abilities of large language models (LLMs). It includes a new social reasoning framework with 25 controls and 5,000 model-generated evaluations.
    }
    
    \item {
        \textbf{Trivial Creative Writing} \citep{wang2024unleashingemergentcognitivesynergy}: This dataset challenges models to generate a coherent story while seamlessly incorporating answers to a set of trivia questions.
    }
    
\end{itemize}

\textbf{Model.}  We selected the Qwen2.5-7B-Instruct \citep{qwen2025qwen25technicalreport} model as the baseline. The models fine-tuned on different numbers of X-chain of thought datasets are used as our comparison models, denoted as $MoR_i, where i = 50, 150, 300, 500$. We believe that after training, the model has acquired MoR capabilities, so simply using the prompt "Let's think step by step." is sufficient to elicit the model’s multi-step reasoning ability. We refer to this prompting strategy as the CoT prompt. For comparison, we also provide a setting where the model is directly instructed to answer the question without any special prompt which is called the IO prompt.

% \begin{figure*}[!t]
%     \centering
%     \includegraphics[width=1\linewidth]{figures/case_study_7.pdf}
%     \caption{Case study comparing the baseline model and $MoR_{150}$ using CoT prompts. The Qwen2.5-7B-instruct model follows the "Let's think step by step." approach but ultimately produces incorrect answers. In contrast, the $MoR_{150}$ model adopts the MoR reasoning method, analyzing problems logically and ultimately arriving at the correct answer.}
%     \label{fig:case_study}
% \end{figure*}

\subsection{Result}
% The summarized results in Table \ref{tab:MoR_result} show that models trained using the MoR approach achieve significant improvements on reasoning tasks. The best performance with the CoT prompt reaches 0.730, surpassing the baseline model by 2.2\%, while the top IO prompt result is 0.734, exceeding the baseline by 13.5\%. 

The summarized results in Table \ref{tab:MoR_result} clearly demonstrate that models trained using the MoR approach achieve substantial and consistent improvements across a wide range of reasoning tasks. Notably, the performance with the Chain-of-Thought (CoT) prompt reaches an accuracy of 0.730, representing a 2.2\% increase over the baseline model, which underscores the effectiveness of structured reasoning in enhancing model capabilities. Interestingly, the highest performance is observed with the Input-Output (IO) prompt, which attains a score of 0.734—exceeding the baseline by a remarkable 13.5\%. This suggests that, while the CoT prompting strategy effectively fosters deeper reasoning, the IO prompts still hold significant value for straightforward tasks.

\subsection{Analysis}
\textbf{Analysis of results. }

 For simple tasks like HotpotQA, most models perform well, with some achieving perfect scores, indicating that basic models are already effective for direct question-answering. However, for complex tasks like StrategyQA and MMLU, MoR models using Chain-of-Thought (CoT) prompts show superior performance, highlighting the importance of structured reasoning chains for complex tasks. The experiments reveal that increasing reasoning templates doesn’t always improve performance, especially with limited training data. The $MoR_{150}$ configuration achieved the optimal chain-of-thought stimulation, and as MoR's chain-of-thought and data grow, explicit guidance may be less necessary, with the IO prompt effectively stimulating reasoning in $MoR_{500}$, achieving a best result of 0.734.

The MoR approach outperforms traditional methods, particularly in multi-step inference and strategy-oriented tasks. While CoT and IO prompts perform similarly, the IO prompt provides a slight advantage in some tasks, showcasing task-specific benefits. These results confirm that integrating MoR training with tailored prompts enhances reasoning abilities, advancing AI in complex problem-solving.

To verify these results, we expanded the test set for both the baseline model and $MoR_{150}$ to 200 samples. As shown in Table X, the extended $MoR_{150}$ maintains a consistent advantage over the baseline.

%但是随着XoT思维链与训练数据的增加，我们认为无需特别的引导，IO prompt也可以一定程度上激发模型的思考能力，例如$XoT_{500}$中，IO prompt设置下取得了最佳的结果0.734。

\textbf{Case study of MoR methods.} 
%As shown in Figure \ref{fig:case_study}, we present an example from the BigTom dataset comparing the baseline model and the $MoR_{150}$ model under the CoT prompt setting. This task primarily assesses LLMs' ability to understand others' mental states and reason about false beliefs. While the baseline model follows the CoT instruction by analyzing the problem step-by-step, it neglects to account for the protagonist's realization that they are wrong and does not consider whether the protagonist's subjective beliefs have changed. Consequently, its reasoning is logically incomplete, which ultimately results in an incorrect answer. In contrast, the model trained with the MoR method leverages its pre-embedded knowledge, selects the most effective reasoning strategy, and applies logical thinking to successfully solve the challenging problem. Notably, MoR approach demonstrates a stronger capacity for theory of mind reasoning, accurately inferring the true beliefs behind the scenario, which highlights its superior understanding of complex mental state reasoning compared to traditional methods. This example highlights the effectiveness of MoR in employing different reasoning strategies to better handle various types of problems.
In Figure \ref{fig:case_study}, we compare the baseline model with $MoR_{150}$ on the BigTom dataset under CoT. This task evaluates LLMs' ability to reason about others' mental states and false beliefs. The baseline model fails to consider the protagonist's changing beliefs, leading to incomplete reasoning and incorrect answers. In contrast, the MoR model selects effective strategies, applying logical thinking to solve the problem correctly. This example demonstrates MoR's strength in theory of mind reasoning, providing superior understanding of complex mental states compared to traditional methods.
\section{Conclusion}
The Mixture of Reasoning (MoR) framework represents a significant advancement in enhancing the reasoning capabilities of large language models by embedding diverse reasoning strategies directly into their parameters. By eliminating the dependency on manually crafted, task-specific prompts, MoR enables LLMs to autonomously select and apply effective reasoning methods tailored to a wide range of complex tasks. Through our two-phase approach—Thought Generation and SFT Dataset Construction—we have demonstrated that MoR not only improves performance over baseline models but also achieves robust generalizability, as evidenced by $MoR_{150}$'s superior results of 0.730 with CoT prompting and 0.734. These findings underscore MoR's potential to redefine how LLMs approach reasoning, offering a scalable and adaptable solution that reduces the burden of prompt engineering. Future work will explore expanding the diversity of reasoning templates and integrating MoR with other advanced training paradigms to further enhance its effectiveness across even more challenging domains.

\bibliography{custom}
\end{document}